\documentclass[conference]{IEEEtran}
\IEEEoverridecommandlockouts
\usepackage{cite}
\usepackage{amsmath,amssymb,amsfonts}
\usepackage{algorithmic}
\usepackage{graphicx}
\usepackage{pgfplots}
\usepackage{textcomp}
\usepackage{booktabs}
\usepackage{xcolor}
\usepackage{multirow}
\usepackage{graphicx}
\usepackage{hyperref}
\pgfplotsset{compat=1.18}

\def\BibTeX{{\rm B\kern-.05em{\sc i\kern-.025em b}\kern-.08em
    T\kern-.1667em\lower.7ex\hbox{E}\kern-.125emX}}
    
\begin{document}

% \title{A Data-Driven Approach for Clotting Prediction in Renal Therapy}
\title{A Data-Driven Approach to Support Clinical Renal Replacement Therapy}

\author{\IEEEauthorblockN{Alice Balboni}
\IEEEauthorblockA{\textit{Department of Applied Science and Technology} \\
\textit{Politecnico di Torino}\\
 Turin, Italy \\
alice.balboni@polito.it}
\and
\IEEEauthorblockN{Luis Escobar}
\IEEEauthorblockA{\textit{Department of Inf.  Eng., Computer Science and Mathematics} \\
\textit{University of L'Aquila}\\
  L'Aquila, Italy \\
luisangel.escobarhernandez@student.univaq.it}
\and
\IEEEauthorblockN{Andrea Manno}
\IEEEauthorblockA{\textit{Department of Inf.  Eng., Computer Science and Mathematics} \\
\textit{University of L'Aquila} \\
 L'Aquila, Italy \\
andrea.manno@univaq.it}
\and
\IEEEauthorblockN{Fabrizio Rossi}
\IEEEauthorblockA{\textit{Department of Inf.  Eng., Computer Science and Mathematics}\\%Information Engineering, Computer Science and Mathematics} \\
\textit{University of L'Aquila}\\
 L'Aquila, Italy \\
fabrizio.rossi@univaq.it}
\and
\IEEEauthorblockN{Maria Cristina Ruffa}
\IEEEauthorblockA{\textit{Department of Applied Science and Technology } \\
\textit{Politecnico di Torino}\\
 Turin, Italy \\
mariacristina.ruffa@polito.it}
\and
\IEEEauthorblockN{Gianluca Villa}
\IEEEauthorblockA{\textit{Dipartimento di Scienze della Salute } \\
\textit{University of Florence}\\
 Florence, Italy \\
gianluca.villa@unifi.it}
\and
\IEEEauthorblockN{Giordano d'Aloisio}
\IEEEauthorblockA{\textit{Department of Inf.  Eng., Computer Science and Mathematics}\\%Information Engineering, Computer Science and Mathematics} \\
\textit{University of L'Aquila}\\
 L'Aquila, Italy \\
giordano.daloisio@univaq.it}
\and
\IEEEauthorblockN{Antonio Consolo}
\IEEEauthorblockA{\textit{Dipartimento di Informatica, Sistemistica e Comunicazione}\\%Information Engineering, Computer Science and Mathematics} \\
\textit{Universita degli Studi di Milano-Bicocca}\\
 Milan, Italy \\
antonio.consolo@unimib.it}

}

\maketitle

\begin{abstract}

\textbf \\
Objective. The study aimed to develop and evaluate the viability of a data driven approach to predict membrane fouling in critically ill patients undergoing CRRT using machine learning algorithms. Moreover, Counterfactual Analysis is used to detect counterfactuals, i.e. the minimal modifications to the input of the machine learning model required to revert a membrane fouling prediction.
\textbf \\
Design. The study utilizes time series from Careggi University Hospital ICU. A subset of 16 specific features was recognized, following the recommendations of clinicians, as the most relevant indicators to train machine learning models for predicting membrane fouling dynamics. To keep the approach simple, interpretable, and amenable to detect reliable counterfactuals, the study mainly focuses on a tabular data approach, not involving the time series’ interdependence inherent within each treatment. Since the number of membrane fouling cases is considerably smaller than the overall number of treatments, the ADASYN oversampling method was utilized as preprocessing step for a more equitable representation of the minority classes. A Shapley values based Counterfactual Analysis is applied to the best prediction model, in order to detect counterfactuals whose quality is measured through a proper score.   
\textbf \\
Results. The specific methods adopted include Random Forest, XGBoost and LightGBM. For all these methods a rebalancing rate of 10\% with respect to the majority class showed the most balance performance with a sensitivity of 0.776 and a specificity of 0.963. The performance obtained by all methods showed to be robust with respect to different length forecasting horizons. The tabular data approach revealed not to be a limitation as it outperformed the Long-Short-Term-Memory Recurrent Neural Networks which intrinsically take into account temporal relationships. It is shown that by reducing the features to 5 via a feature selection method, we obtain more simple and interpretable models, without compromising too much the accuracy.  The adopted Counterfactual Analysis method is able to detect  counterfactuals which seems promising according to the considered quality score. %Across the evaluation of the different lag values the performance of the model is does not impact the overall effectiveness of the predictive algorithms suggesting that the models are robust enough to maintain their predictive capabilities regardless of the lag configuration used.
\textbf \\
Conclusions. The experimental study provides promising results concerning the adoption of data-driven machine learning methods to predict membrane fouling events during CRRT. The practical implications for clinicians and nurses managing CRRT are significant; additionally, the interpretability afforded by using a reduced set of features enhances the understanding on how the model arrives at their conclusions without sacrifying too much predictive power. The predictions of the model and the associated Counterfactual Analysis can inform therapeutic adjustments leading to more effective patient care while minimizing the risk of membrane fouling.   %The adoption of more sophisticated modeling techniques, such as recurrent neural networks and/or long short-term memory networks can significantly improve prediction accuracy while effectively maintaining the intricate time-related patterns. 

\end{abstract}


\section{Introduction}\label{sec:intro}
Continuous Renal Replacement Therapy (CRRT) is a crucial therapeutic approach widely employed in Intensive Care Units (ICUs) to support patients suffering from Acute Kidney Injury (AKI) and other severe, multifactorial health conditions (e.g., \cite{villa2025use}). This modality is particularly beneficial for hemodynamically unstable patients, as it employs a semipermeable filter to enable slow and controlled removal of fluids and accumulated toxins and solutes \cite{herrera2024continuous}. Although the nominal operational lifespan of a CRRT filter is approximately 72 hours, clinical evidence indicates that therapy is prematurely discontinued in 75.7\% of patients due to unanticipated membrane fouling \cite{li2021effect}. Filter clotting represents the primary cause of premature filter lifespan reduction during CRRT. This phenomenon is defined as the formation of thrombi within the semipermeable membrane and extracorporeal circuit, resulting from activation of the coagulation cascade due to blood-membrane interactions, shear stress, and inadequate anticoagulation. Beyond accelerating filter and circuit wear, this complication causes CRRT under-dosing and delays renal function recovery, with attendant adverse clinical sequelae \cite{tsujimoto2022prolong}. \par
There is an urgent clinical need for accurate risk assessment tools to enable timely detection of filter wear and therapy inadequacy by healthcare providers. Early prediction of such complications facilitates proactive interventions, mitigating adverse outcomes and optimizing patient prognosis.

Deep learning models have demonstrated notable effectiveness in managing the complexities of high-dimensional healthcare data (see e.g., \cite{leung2024deep,jiao2020deep,norgeot2019call}). They excel at learning meaningful representations of intricate medical concepts and capturing nonlinear interactions that are often present in clinical datasets. However, they come with their own set of drawbacks, such as a complex training phase due to essentially vanishing/exploding gradient phenomena (see e.g., \cite{hochreiter1998vanishing,lecun2015deep}), and challenges related to interpretability, making it difficult for researchers and practitioners to understand how decisions are made by these complex models. \par 

In contrast, shallow machine learning methods applied to tabular data, have demonstrated potential equivalence to deep networks in specific contexts (e.g., \cite{manno2023comparing,maqsood2022survey}), often providing clearer insights into the decision-making process while avoiding some of the complexities associated with deeper architectures \cite{bermeitinger2023make}. Specifically, ensemble methods \cite{dietterich2000ensemble} can enhance the performance of traditional machine learning by combining the predictions of multiple models (e.g., \cite{gao2021improving,manno2024ensemble}). By integrating the outputs of various shallow techniques, ensemble methods can mitigate individual model biases and improve overall accuracy. This approach not only strengthens the robustness of predictions but also maintains the interpretability of the foundational models, allowing for a more comprehensive understanding of the decision-making process. \par 
Consequently, the primary goal of this study is to develop and evaluate the viability of a data-driven approach for the prediction of membrane fouling events in critically ill patients undergoing CRRT using machine learning algorithms with a focus on tabular data and ensemble methods. In particular Random Forest \cite{breiman2001random}, XGBoost \cite{chen2016xgboost}, and LightGBM \cite{ke2017lightgbm} are compared in the prediction of clotting events, as they have been successfully adopted in a variety of healthcare applications (e.g., \cite{gurm2014random,zelli2023classification,liao2022lightgbm}). Differently from the interesting work in \cite{yang2024development}, where the incidence of a potential early clotting event (within 48 hours) is predicted via a binary logistic regression model exploiting variables collected at the beginning of the treatment (including patient demographic and clinical conditions), here we are interested in predicting membrane fouling events with a certain advance by continuously observing the status of the signals in the CRRT machine. The adopted tabular data approach, besides experimentally showing to outperform Long-Short-Term-Memory (LSTM) Recurrent Neural Networks (RNNs) \cite{hochreiter1997long} (directly processing time series data), is amenable to Counterfactual Analysis \cite{derczynski2016complementarity} , which suggests the way to revert an adverse prediction and, potentially, avoid a costly clotting event. A model agnostic Counterfactual Analysis method based on Shapley values (see \cite{albini2022counterfactual}) is applied to the best performing membrane fouling prediction model, with promising results.\par

The remainder of the paper is organized as follows: in Section \ref{sec:methods} we present the overall methodology, including the description of the dataset, of the machine learning binary classification task, of the data pre-processing techniques, and of the adopted prediction models. In Section \ref{sec:results} the experiments on the predictive models and the obtained results are reported and analysed together with the Counterfactual Analysis. Section \ref{sec:conclusions} is devoted to concluding remarks.

\section{Methods}\label{sec:methods}

\subsection{The dataset and the labeling process}\label{subsec:dataset}
The dataset for the study consists of CRRT time series from the Careggi University Hospital ICU, with records collected over an extended period from 2011 to 2014. Patient-specific information is systematically documented every minute, leading to the generation of approximately 1,440 data points for each individual treatment. A total of 796 treatments were provided, from which a subset of 91 cases was identified as treatments in which clotting occurred at some point. The remaining 705 treatments correspond to therapies that did not exhibit any signs of clotting at all throughout their duration. The dataset consists of 16 features that can be divided into four distinct categories: treatment configuration, machine preassure, prescription variables, and critical clotting variables as shown in Table \ref{tab:features_table}. Each category encompasses a unique set of factors that contribute to the overall understanding of clotting dynamics, potentially enabling the model to predict the likelihood of clot formation under various conditions.

\begin{table}[h!]
\centering
\begin{tabular}{|p{0.08\textwidth}|p{0.28\textwidth}|p{0.05\textwidth}|}
\hline
\textbf{Category} & \textbf{Feature description} & \textbf{Name}\\ \hline
\multirow{4}{0.08\textwidth}{treatment configuration}
    & time stamp & ts \\ \cline{2-3}
    & serial number identification for the treatment & id \\ \cline{2-3}
    & patient weight & w \\ \cline{2-3}
    & filter set used for the treatment & set	 \\ \hline
\multirow{4}{0.08\textwidth}{machine pressure} 
    & pressure created by pulling blood from patient & P\_acc \\ \cline{2-3}
    & pressure to push blood through filter &  P\_filt \\ \cline{2-3}
    & pressure measured in effluent line & P\_eff\\ \cline{2-3}
    & pressure created by returning blood to patient & P\_ret \\ \hline
\multirow{5}{0.08\textwidth}{prescription variables} 
    & blood flow set by the blood pump & Q\_bp \\ \cline{2-3}
    & replacement fluid flow rate & Q\_rep \\ \cline{2-3}
    & dialysate fluid flow rate &	Q\_dial \\ \cline{2-3}
    & pre filter replacement fluid flow rate & Q\_pbp \\ \cline{2-2}
    & net ultrafiltration flow rate		& Q\_pfr \\ \hline
\multirow{3}{0.08\textwidth}{critical clotting} 
    & difference between return and filter pressures & $\Delta$P				\\ \cline{2-3}
    & pressure gradient across the filter membrane & TMP \\ \cline{2-3}
    & access transmembrane pressure for 
    therapeutic plasma exchange treatment & TMPa\\ \hline
\end{tabular}
\caption{Feature Variables.}
\label{tab:features_table}
\end{table}
In this work we adopt a supervised learning binary classification approach, requiring that each data entry associated to a specific time stamp, is classified through the presence (positive class) or absence (negative class) of the clotting/clogging event (from now on referred to as blocking) at time ts+$\delta$. Despite
the available dataset does not explicitly contain such information, the data labeling is easily performed based on the values of  features $\Delta\mbox{P}$, and the pair $\mbox{P}_{\mbox{fil}}$-$\mbox{TMP}$, quantifying fiber clotting and membrane clogging, respectively. 
Two specific rules are taken into consideration in order to accurately label and identify the various membrane fouling events that may arise throughout the therapy.
% $DeltaP$ is defined by Equation \ref{eq:DeltaP}, which illustrates the relationship between inlet and outlet pressures during filtration. 

% \small
% \begin{equation}
% DeltaP = Pressure_{filter} - Pressure_{return} -25mmHg
% \label{eq:DeltaP}
% \end{equation}
% \normalsize

% On the other hand, $TMP$ is obtained from Equation \ref{eq:TMP}, which provides an essential measure of the driving force of filtration across the membrane. 

% \begin{equation}
% %TMP = [\frac{(Pressure_{filter} + Pressure_{return}) / 2]} - Pressure_{effluent}
% %TMP = \left[\frac{Pressure_{{filter}} + Pressure_{{return}}}{2}\right] - %Pressure_{{effluent}}
% \resizebox{0.8\columnwidth}{!}{$TMP = \left[\frac{Pressure_{{filter}} + Pressure_{{return}}}{2}\right] - Pressure_{{effluent}}$}
% \label{eq:TMP}
% \end{equation}
 The first rule is based on a reference value of $\Delta\mbox{P}$ related to the beginning of the treatment, and denoted as $\Delta\mbox{P}_{\mbox{ref}}$.
 To enhance the robustness of the labeling process, $\Delta\mbox{P}_{\mbox{ref}}$ is  computed as the median of $\Delta\mbox{P}$ over the initial ten minutes of the treatment. 
 According to the first rule, a transition to the positive class occurs when $\Delta\mbox{P}$ exceeds $\Delta\mbox{P}_{\mbox{ref}}$ by at least $100$ mmHg. Formally, \begin{equation}
\Delta\mbox{P} > \Delta\mbox{P}_{\mbox{ref}} + 100 \mbox{ mmHg}.
\label{eq:Condition1}
\end{equation}
This indicates that the process encounters a substantial level of resistance, strongly suggesting the possible onset of clot formation.  If condition (\ref{eq:Condition1}) is not met, the second rule triggers a transition to the positive class when two different alarms are activated. The first precautionary alarm, which is related to potential coagulation, is activated when 
\begin{equation}
\mbox{P\_{filt}} \le 450 \mbox{ mmHg}.
\label{eq:Coagulation_in_progress}
\end{equation}
 % to a potential activated when to indicates that the concurrent activation of the thresholds for both the informative alarm and the precautionary alarm serve as an indicator of a possible incident. An event in progress within the filter occurs when the condition specified in Equation \ref{eq:Coagulation_in_progress} is fulfilled. 
 The second alarm related to excessive TMP is triggered when \begin{equation}
\mbox{TMP} > \mbox{TMP}_{\mbox{filter}}
\label{eq:Excessive TMP}
\end{equation} where the $\mbox{TMP}_{\mbox{filter}}$ value is specific of the adopted filter. The values $\mbox{TMP}_{\mbox{filter}}$ for the considered filters can be found in Table \ref{tab:filter_table}.

\begin{table}[h!]
\centering
\begin{tabular}{|c|c|}
\hline
\textbf{Filter Code} & \textbf{TMP filter (mmHg)} \\ \hline
septeX              & 450                       \\ \hline
ST150               & 450                       \\ \hline
Xiris               & 450                       \\ \hline
TPE2000             & 500                       \\ \hline
X-MARS              & 600                       \\ \hline
M100                & 450                       \\ \hline
ST60                & 450                       \\ \hline
HF1000              & 500                       \\ \hline
ST100               & 450                       \\ \hline
HF1400              & 500                       \\ \hline
\end{tabular}
\caption{Filter codes and their TMP filter values.}
\label{tab:filter_table}
\end{table}

\begin{figure}[htbp]
    \centering
    \includegraphics[scale=0.65]{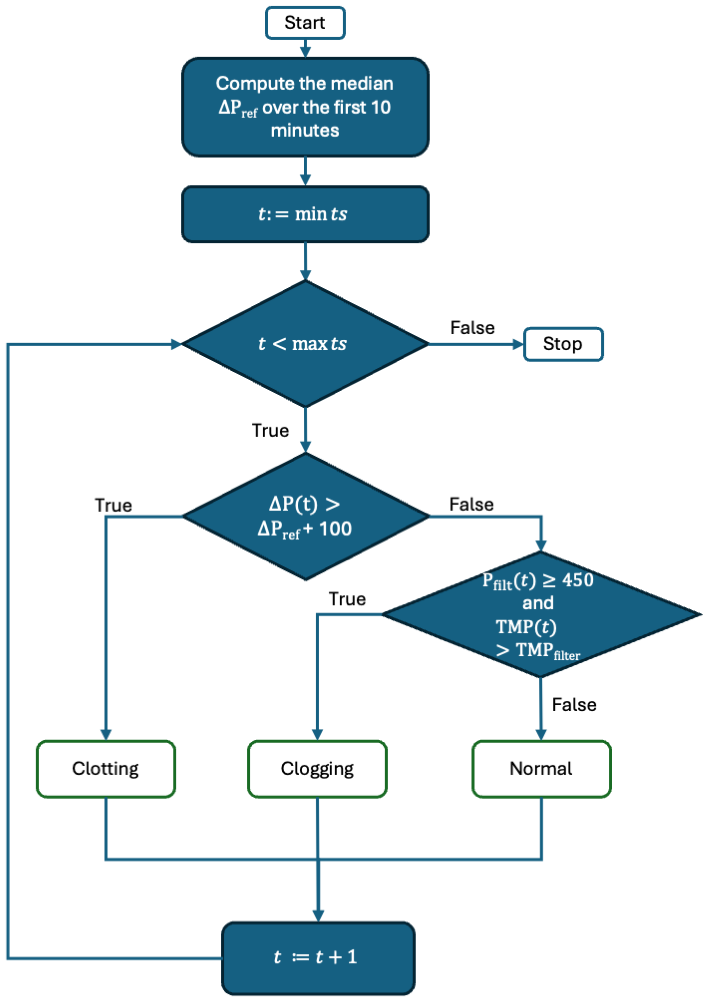}
    \caption{labeling process.}
    \label{fig:Clotting label}
\end{figure}

%To mitigate the identification of false positives due to the presence of noise or external factors, clinicians have recommended that a minimum duration of 15 minutes of continuous conditions must be satisfied in order to classify a potential event as one, thereby reducing the likelihood of misclassifying a source of noise as an actual clotting event. 
Figure \ref{fig:Clotting label} illustrates the decision flow used to categorize a sample as an event or not.

\subsection{Tabular data design}\label{sec:static_data}
Despite all features in the original dataset are organized as time series, in this work we focus on a tabular data approach. In particular, the training set is organized so that the input associated with each training sample is the set of feature values at a single time stamp, ignoring the interdependence between distinct treatments and the interdependence inherent within each treatment. Therefore the feature "ts" were not explicitly included. Moreover, for a more general approach, even the features related to the patient ("id" and "w") and to the filter set used ("set") were excluded, resulting in a final number of 12 features. Then, each single set of features associated with a generic time stamp $\mbox{ts}$ is labeled with the binary outcome computed (as in Section \ref{subsec:dataset}) for the time instant $\mbox{ts}+\delta$ where $\delta$ is the desired forecast horizon.\par
The tabular data choice is due to the following motivations:
\begin{itemize}
    \item Time series require the use of complex and demanding learning architectures, such as RNNs. In contrast, using tabular data leads to a more manageable and interpretable traditional supervised learning problem that incorporates 12 distinct features to predict the target blocking outcome;
    \item Tabular data enhances the ability to utilize powerful explainability tools like Shapley values (see \cite{lundberg2017unified}), providing insights into the relationship between features and the blocking phenomenon;
    \item Despite ignoring the temporal relationships and the treatment and patient dependence may result in a potential loss of information, the tabular data approach allows to better exploit the size of the dataset, which consists of a relatively limited amount of treatments and patients, but of a large amount of single time stamps samples;
    \item Tabular data are suited for a straightforward application of Counterfactual Analysis (CA) methods (see Section \ref{sec:counter}), which determine interesting guidelines to reverse adverse predictions.
\end{itemize}
Noticeably, as reported in Section \ref{sec:results}, the tabular data methods outperform in terms of predictive accuracy the time series powered LSTM RNNs. \par

%The study leverages a dataset during the training phase that ignores the independence between distinct treatments as well as the interdependence inherent within each treatment; rather, each data point corresponding to a particular temporal instant is regarded as a static, time-invariant instance. This methodological choice simplifies the analysis, transforming what would typically be a complex time series analysis into a more manageable supervised learning problem that incorporates 17 distinct features to predict the desired clotting outcome.

\subsection{Rebalancing methods}
Since in the dataset the number of treatments affected by blocking is much smaller than the overall number of treatments (see Section \ref{subsec:dataset}), and since any blocking events typically occurr at the end of relatively long treatments, the tabular training set is highly affected by class imbalance. For this reason, the Synthetic Minority Oversampling Technique (SMOTE) \cite{chawla2002smote} was utilized as  preprocessing step to allow for a more equitable representation of minority classes potentially improving the performance of machine learning models. This step means that synthetic data was generated temporarily during the training process and was subsequently eliminated once it fulfilled its intended function. In particular, the Adaptive Synthetic Sampling method (ADASYN) \cite{he2008adasyn} that enhances the SMOTE algorithm by concentrating on the generation of synthetic samples for minority class instances that present greater classification challenges, enhancing the model’s proficiency by learning from these cases and increasing overall predictive accuracy. To ensure a balanced representation of the classes and mitigate potential overfitting and biases, different balancing percentages were tested to identify the optimal ratio that maximizes the models' performance without compromising its ability to generalize.

\subsection{Model selection and hyperparameters configuration}
A crucial aspect is related to the selection of the machine learning model to obtain good predictive performance but also to ensure simplicity and resilience against overfitting. In light of this, ensemble methods emerge as a particularly advantageous approach, as these techniques skillfully combine the predictions from several models to enhance accuracy and generalization. Different studies (see Section \ref{sec:intro}) have shown that ensemble methods can outperform traditional methods in predicting clinical outcomes, including disease progression and treatment responses, reducing variance and bias, making them ideal for complex clinical scenarios where individual model predictions may vary widely. Because of this, the specific techniques used included Random Forest, XGBoost, and LightGBM.\par 
We also report comparisons with LSTM  which are state-of-the-art RNNs widely use in healthcare contexts (see e.g., \cite{hou2019lstm,men2021multi,srikantamurthy2023classification}). Such comparisons are used to show that in the investigated case-study the tabular data approach, in spite of the aforementioned benefits, does not actually yield accuracy losses with respect to more sophisticated methods which directly process time series.\par

The calibration of the hyperparameters (model configuration parameters that are not tuned during the training phase) has a strong impact on the performance of machine learning models. Concerning the considered ensemble methods, the hyperparameters have been tuned via the well-known 10-fold grid-search cross-validation (e.g., \cite{bishop2006pattern}), which is a simple heuristic search to determine promising hyperparameter values.\par 

Due to a more sophisticated architecture, the hyperparameter tuning of LSTM can be much more critical for the performance. Therefore, we have optimized the LSTM hyperparameters using Optuna \cite{akiba2019optuna} which is an advanced framework that implements a Tree-structured Parzen Estimator algorithm \cite{watanabe2023tree} to identify the best hyperparameters that optimise a given fitness score. Since, as shown in the Section \ref{sec:results}, we focus on the Sensitivity and Specificity metrics, we have chosen the Time Window F1 Score, which is defined as the harmonic mean between Sensitivity and Specificity \cite{derczynski2016complementarity}, as the objective score for Optuna.

\section{Results}\label{sec:results}

\subsection{Post-processing function}
The expected behavior of the prediction model should be like a step function. Indeed, in a first phase corresponding to the beginning of the treatment, the target variable is fixed to zero, i.e. all samples belong to the negative class (no blocking events). %meaning that, at the beginning of the treatment, no blocking events happen. 
However, as the treatment progresses, depending on the conditions of the patient and of the filter membrane, a blocking event may occur, resulting in a transition from the negative class to the positive class. In preliminary experiments, the predictions provided by the trained model often showed a slight oscillatory pattern in the transition from negative to positive class. Then, after a certain number of time stamps, the predictions correctly and definitively turn to the positive class. %In preliminary experiments the predictions provided by the trained model often showed a slight oscillatory pattern in the transition from negative to positive class. Than, after a certain amount of time stamps, the predictions correctly turn definitely to the positive class.  %See Figure \ref{fig:Predictor}.

\begin{figure}[htbp]\label{fig:profiles}
    \centering
       \includegraphics[scale=0.5]{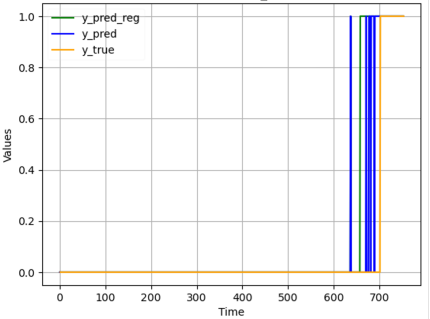}
    \caption{The real (in yellow), predicted (in blue), and post-processed predicted (in green) profiles of the class labels, for an exemplificative instance.}
    \label{fig:Predictor}
\end{figure}

Although the oscillatory phenomenon is unrealistic from a practical point of view, it does not constitute a limitation from a predictive perspective as it often occurs in correspondence with a real initial blocking status, suggesting that the features at that point are critical indicators of a potential blocking event.
%While this oscillatory phenomenon is unrealistic from a practical point of view, it often occurs before the actual blocking event, indicating that the features at that point are critical indicators of a potential blocking event. 
%From a clinical point of view, this phenomenon is not particularly bad, 
In other words, the model is capable of generating a potential alert attributable to the oscillatory dynamics; such a response might be enough to modify the therapeutic approach and mitigate future complications, even in instances where the signal does not conform to an ideal step function. For this reason, the oscillation is interpreted as the indicative signal of a blocking alarm if it occurs frequently within a relatively brief temporal interval. Consequently, a regularized (filtered) post-processed prediction function is generated in the form of a step function that can be utilized to evaluate the model in a more fair way. In particular, by observing a time window of $d$ consecutive time stamps, if the percentage of time stamps in which predictions switch between the two classes is larger than a certain threshold $\epsilon d$ with $\epsilon \in (0,1)$, then the oscillation can be considered as an indicator of the initial blocking status and in the post-processed prediction all the oscillating time stamps (and all subsequent ones) are automatically set to the positive class. Otherwise, the oscillation is associated with irrelevant noise and in the post-processed prediction all the oscillating time stamps are set to the negative class. %This can be viewed as a post-processing filter for the output produced by the machine learning model. 
\par 
Such behavior is depicted in Figure \ref{fig:profiles}, where the profiles of the real class labels (yellow), of the predictions generated by the machine learning model (blue), and of the post-processed predictions (green) are represented for a %exemplificative 
representative instance. In a long first phase the three profiles overlap at the zero value associated to the negative class. Around time stamp 630 the predicted blue profile has a first quick oscillation, then it returns fixed to zero and around time stamp 670 it starts a long oscillation phase between the two classes. The yellow real class labels profile turns to the positive class around time stamp 700. According to the mechanism described above, the green post-processed prediction profile ignores the first oscillation of the blue profile maintaining the prediction to the negative class, then interprets the second oscillation phase at time stamp 670 as an indicator of the initial blocking event and is stably set to one, slightly anticipating the real blocking event.

%is determined the prediction model produces an oscillation between 
%In particular, as depicted in Figure \ref{fig:Window}, if the model produce an oscillation between positive and negative class of certain duration, say $x$, the regularized prediction is set to positive class definitely. 
%To do this, a tolerance window was implemented. See Figure \ref{fig:Window}. The concept underlying this is that a forecast at time t-x, where t represents the true blocking point and x represents a delta value, could be a useful blocking alert if the value of x is not too large.

% \begin{figure}[htbp]
%     \centering
%     \includegraphics[width=\columnwidth]{Images/4.2_predictor_reg.png}
%     \caption{Example of a figure caption.}
%     \label{fig:Window}
% \end{figure}

\subsection{Experiments' configuration}
%The different ensemble methods already incorporated a parameter aimed at tackling the imbalance issue; nonetheless, the implications of additionally implementing a balancing technique were unknown.
Concerning the tabular data methods, four distinct scenarios were examined; the first consists in training the models without any balancing alterations, thereby maintaining the original distribution between the two categories. In the second, a balance of 5\% was established, i.e. the minority class was augmented through synthetic samples to constitute 5\% of the majority class. In the third and forth scenarios the minority class representation was raised to 10\% and 20\% respectively. \par  
For all the methods, different forecasting horizons (lags) were considered, from 10 minutes to 60 minutes with a 10-minute granularity.

\subsection{Models' predictive performance}
The models' performance are measured in terms of Sensitivity and Specificity with respect to the positive class.\par 
First, Table \ref{tab:performance} shows the outcomes of the three ensemble models trained on tabular data for different balancing values on a 10-minute lag using the ADASYN method. Due to the computational cost of Random Forest, the experiments for 5\%, 10\%, and 20\% were not included but are compared later, using the rebalancing percentage with the best results.

\begin{table}[h]
    \centering
    \begin{tabular}{llcc}
        \toprule
        Model & ADASYN & Sensitivity & Specificity \\
        \midrule
        {Random Forest} & None & 0.446 & 0.994 \\
        \midrule
        {XGBoost} & None & 0.304 & 0.999 \\
        & 5\% & 0.609 & 0.971 \\
        & 10\% & 0.776 & 0.963 \\
        & 20\% & 0.806 & 0.884 \\
        \midrule
        {LightGBM} & None & 0.846 & 0.621 \\
        & 5\% & 0.667 & 0.969 \\
        & 10\% & 0.682 & 0.948 \\
        & 20\% & 0.677 & 0.966 \\
        \bottomrule
    \end{tabular}
    \caption{Models' Comparison with different rebalancing levels.}
    \label{tab:performance}
\end{table}
The findings indicate that the ADASYN sampling technique significantly enhanced the performance of the models. A sampling rate of 10\% demonstrates the most balanced performance, suggesting that this percentage of synthetic data effectively boosts the capacity of the model to detect positive instances while keeping the false positive rate minimal. Consequently, Table \ref{tab:performance_minutes} explores the performance of all the compared models for different lag values adopting a balancing value of 10\% for the ensemble methods.

\begin{table}[h]
    \centering
        \caption{Models' comparison at different forecasting horizons.}
    \label{tab:performance_minutes}
    \begin{tabular}{lccc}
        \toprule
        Model & Minutes & Sensitivity & Specificity \\
        \midrule
        {Random Forest} 
        & 10 & 0.720 & 0.965 \\
        & 20 & 0.674 & 0.965 \\
        & 30 & 0.651 & 0.966 \\
        & 40 & 0.714 & 0.964 \\
        & 50 & 0.670 & 0.966 \\
        & 60 & 0.681 & 0.966 \\
        \midrule
        {XGBoost} 
        & 10 & 0.776 & 0.963 \\
        & 20 & 0.744 & 0.959 \\
        & 30 & 0.744 & 0.963 \\
        & 40 & 0.844 & 0.902 \\
        & 50 & 0.736 & 0.884 \\
        & 60 & 0.741 & 0.964 \\
        \midrule
        {LightGBM} 
        & 10 & 0.682 & 0.948 \\
        & 20 & 0.681 & 0.964 \\
        & 30 & 0.759 & 0.834 \\
        & 40 & 0.792 & 0.963 \\
        & 50 & 0.750 & 0.967 \\
        & 60 & 0.834 & 0.965 \\
        \midrule
        {LSTM} & 10 & 0.595 & 0.951\\
 &  20 & 0.583 & 0.967  \\
 &  30 & 0.588 & 0.954  \\
 &  40 & 0.511 & 0.953 \\
 &  50 & 0.499 & 0.953  \\
 &  60 & 0.450 & 0.947 \\
        \bottomrule
    \end{tabular}
\end{table}

Across the different lag values, the performance of the models is similar, indicating that the  lag does not significantly impact comparisons and that models are robust with respect to the tested lags. In terms of Specificity, the ensemble methods on tabular data have comparable performance with respect to LSTM, however the latter show substantially lower accuracy in terms of Sensitivity. Therefore, overall the tabular data approaches perform better, with XGBoost showing to be the best ensemble model (slightly better than LightGBM).\par 
To assess the ways in which the different features affect the prediction process within the models, we applied the SHapley Additive exPlanations (SHAP) method to the best performing model, i.e. XGBoost. The SHAP method is a valuable explainability tool based on Shapley values for analyzing the impact of each feature on the predictions produced by a machine learning model, offering clarity on which specific variables are influencing the results and how they contribute to the overall prediction process. Shapley values, originating from game theory, tell how much a feature influences positively (toward the positive class) or negatively (toward the negative class) the final prediction of the trained model. The Shapley value of a feature is determined by observing the output of the model retrained without that feature. The absence of the feature is simulated by using a reference value taken from a background distribution $D$ of points \cite{strumbelj2010efficient} (typically the whole training set or samples from the opposite class). \par
\begin{figure}[htbp]
    \centering
    \includegraphics[width=\columnwidth]{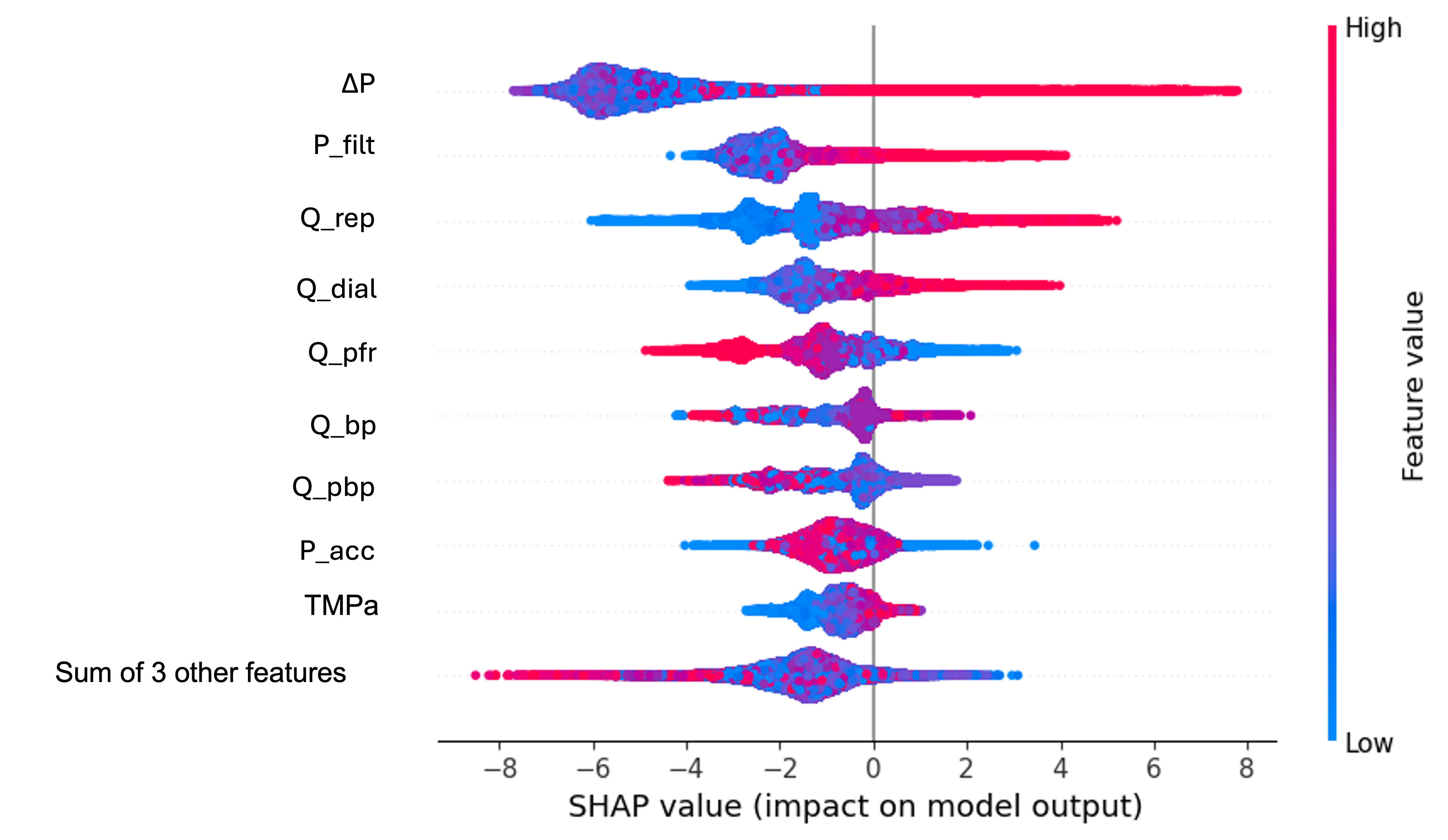}
    \caption{SHAP analysis for the XGBoost model with 10 minutes lag.}
    \label{fig:SHAP values}
\end{figure}

 
Figure \ref{fig:SHAP values} illustrates the SHAP analysis for the XGBoost model with a 10-minute lag, showing the contribution of each feature to the model predictions and highlighting the key factors driving the decision-making process. In the figure, the features are ranked (from top to bottom) according to their importance, and the horizontal axis represents the Shapley value, i.e., the magnitude of each feature’s positive or negative contribution to the model output. The color scale indicates high (red) and low (blue) feature values. The violin plots illustrate the distribution of samples for each feature. \\
\indent Table \ref{tab:performance_minutes_top_5} shows the performance of XGBoost using the five most significant features across the different lag values. Comparing such results with those of Table \ref{tab:performance_minutes}, it is evident that the adoption of a simpler and more interpretable reduced set of features comes with a relatively small accuracy loss.

\begin{table}[h]
    \centering
        \caption{Models' comparison at different forecasting horizons for the top 5 features.}
    \label{tab:performance_minutes_top_5}
    \begin{tabular}{lccc}
        \toprule
        Model & Minutes & Sensitivity & Specificity \\
        \midrule
        {XGBoost} 
        & 10 & 0.765 & 0.919 \\
        & 20 & 0.678 & 0.948 \\
        & 30 & 0.681 & 0.950 \\
        & 40 & 0.693 & 0.947 \\
        & 50 & 0.676 & 0.945 \\
        & 60 & 0.665 & 0.948 \\
        \bottomrule
    \end{tabular}

\end{table}

\subsection{Counterfactual Analysis}\label{sec:counter}

Given a machine learning model and a reference input feature vector  $x \in \Re^n$ for which the model yields a positive prediction, Counterfactual Analysis (CA) is an explainability technique that determines a feature vector $x'$, referred to as a Counterfactual Explanation (CE), that is as close as possible to $x$ and for which the prediction is reversed to negative. \par % class. \par
%Given a machine learning model and a reference input feature vector $x \in \Re^n$ for which the model yields a positive prediction, CA is an explainability technique determines the feature vector $x'$, denoted as Counterfactual Explanation (CE), as close as possible to $x$ for which the prediction is reverted to negative. \par 
In the investigated case-study, CA can suggest useful guidelines to determine the minimal prescription modifications required to restore the regular functioning of the CRRT machine in the presence of a blocking prediction. Indeed, the principle of proximity between the reference vector $x$ and the CE $x'$ is in line with the natural tendency of clinicians to not modify too much the therapy they have carefully set for the patient.\par 
CEs are typically determined by solving Mathematical Optimization problems embedding the structure of the machine learning predictive model, hence CA is often applied for simple or generalized linear models (see e.g., \cite{magagnini2024nearest,artelt2019computation,guidotti2024counterfactual}). However, in our application we are interested in CA for the best performing model XGBoost, which cannot be directly embedded in an optimization problem. Therefore, we adopt the model agnostic Shapley values-based strategy proposed in \cite{albini2022counterfactual}. Without entering into details that can be found in \cite{albini2022counterfactual}, the idea is to use the K-nearest-neighbors of $x$ belonging to the opposite class as background distribution to compute the Shapley values. Then, such Shapley values are used to construct a geometric direction along which the CE $x'$ is determined. Indeed, exploiting the local information from the K-nearest-neighbors is shown to be an effective way to find good quality CEs that are closer to $x$. \par 
An important aspect is the way to evaluate the quality of CEs. We consider the {\it counterfactual ability} score, representing the cost $c(x,x')$ of moving from $x$ to $x'$ (the smaller is $c(x,x')$ the better is $x'$). Given the geometrical nature of this counterfactual approach, a key factor is the metric used to compute the distance between feature vectors. Indeed, an unbalanced scale of feature ranges may affect the computations of the counterfactuals. %\red{While in \cite{shapcounter} a quantile shift data transformation is combined with an $\ell_1$ (Manhattan) distance, i.e. the cost is defined as $c_{\ell_1}(x,x')=\|Q(x)-Q(x')\|_1$ where $Q(\cdot)$ is the quantile transformation, here we compare such choice with the Mahalanobis distance $c_M(x,x')=\sqrt{(x-x')^T \sum^{-1}(x-x')}$, where $\sum$ is the covariance matrix... Using an $\ell_1$ cost on quantile-transformed features provides a rank-based, marginally comparable and additive distance that often encourages sparse counterfactuals, but it neglects correlations across features. The Mahalanobis distance measures changes in standard-deviation units while accounting for the data covariance, penalizing shifts in low-variance directions and promoting correlated moves, thus yielding more coherent and plausible counterfactuals for correlated clinical variables.}
While in \cite{albini2022counterfactual} a quantile-shift data transformation is combined with an $\ell_1$ (Manhattan) distance, i.e. the cost is defined as $c_{\ell_1}(x,x')=\|Q(x)-Q(x')\|_1$ where $Q(\cdot)$ is the quantile transformation, here we compare this choice with the Mahalanobis distance $c_M(x,x')=\sqrt{(x-x')^T \Sigma^{-1}(x-x')}$, where $\Sigma$ is the covariance matrix. Using an $\ell_1$ cost on quantile-transformed features yields a rank-based distance that makes marginal changes more comparable across variables and, due to its additivity, often encourages sparse counterfactuals; however, it ignores dependencies and correlations among features. In contrast, the Mahalanobis distance measures changes in standard-deviation units while accounting for the data covariance, penalizing shifts along low-variance directions and allowing correlated moves, which tends to produce more coherent and plausible counterfactuals, especially with correlated clinical variables (see e.g., \cite{penny1999multivariate,pokrajac2005applying,guerrero2022mahalanobis}).

\par
In our experiments we have computed counterfactuals for 1000 testing feature vectors that have been correctly classified as positive by the XGBoost model with 60 minutes lag. On the one hand, as done in \cite{albini2022counterfactual}, the experiments aim to evaluate the benefits of using the local information of the set of 10-nearest-neighbors of opposite class (referred to as $D_{10-NN}$), with respect to other standard and more global choices for the Shapley values background set, i.e. the whole training set ($D_{TRAIN}$), the samples of the training set actually of opposite class with respect to the reference $x$ ($D_{D-LAB}$), and the samples of the training set correctly predicted of opposite class from XGBoost ($D_{D-PRED}$). On the other hand, we want to compare the $c_{\ell_1}$ with the proposed $c_M$.\par
\begin{figure}[h!]

%\begin{subfigure}
    \centering
    \includegraphics[scale=0.25]{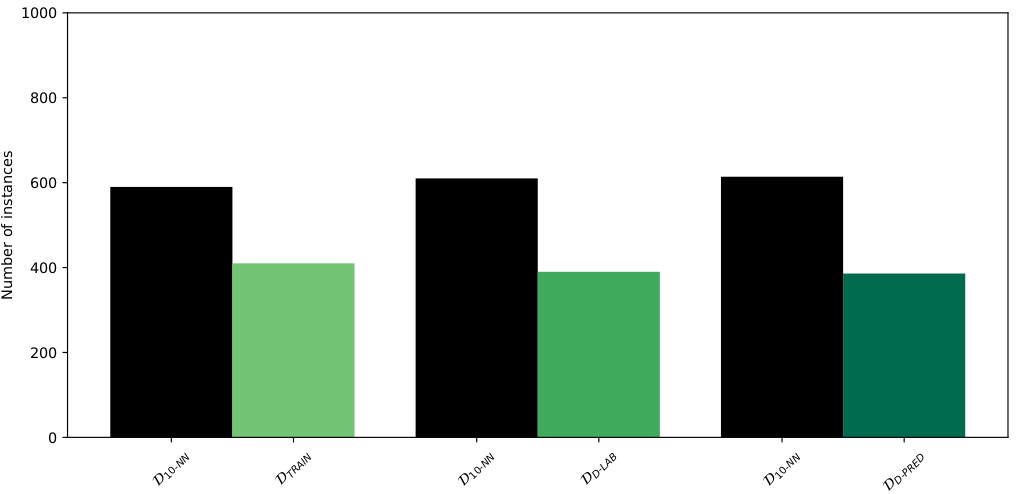}
    \caption{Counterfactual ability-$c_{L_1}$: number of successes over 1000 tests.}
    \label{fig:cl1}
%\end{subfigure}
%\begin{subfigure}
    \centering
    \includegraphics[scale=0.25]{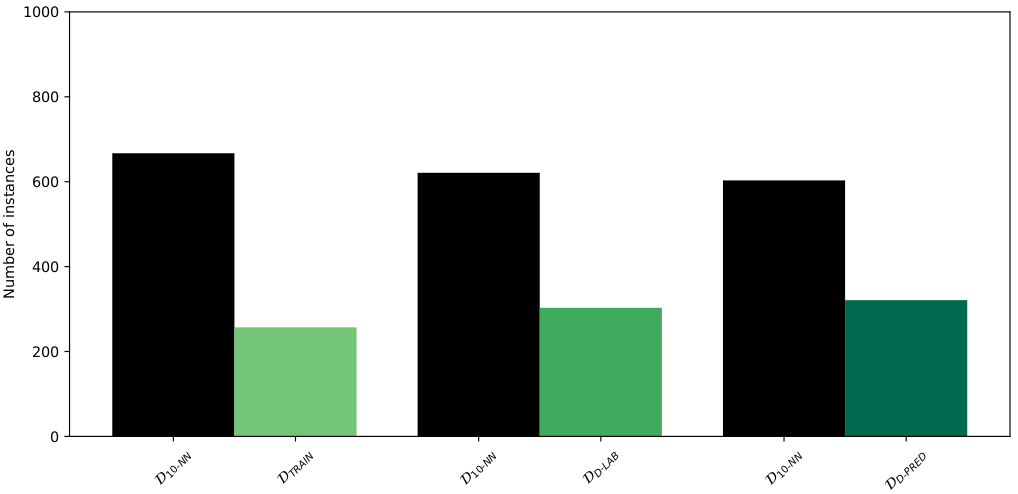}
    \caption{Counterfactual ability-$c_{M}$: number of successes over 1000 tests.}
    \label{fig:cm}
%\end{subfigure}
\end{figure}

Figures \ref{fig:cl1} and \ref{fig:cm} report, for the cost functions $c_{L_1}$ and $c_M$, respectively, the number of times (out of 1000 test instances) in which the counterfactual explanations obtained using the $D_{10\text{-}NN}$ Shapley value background outperform those computed with $D_{D\text{-}TRAIN}$, $D_{D\text{-}LAB}$, and $D_{D\text{-}PRED}$ (black bars), and vice versa (green bars). A comparison of the two figures shows that counterfactuals based on $D_{10\text{-}NN}$ consistently outperform those derived from $D_{D\text{-}TRAIN}$, $D_{D\text{-}LAB}$, and $D_{D\text{-}PRED}$, with the superiority being more pronounced when the proposed distance $c_M$ is adopted (Figure \ref{fig:cm}).
%Figures \ref{fig:cl1} and \ref{fig:cm} reports, for respectively $c_{L_1}$ and $c_M$, the number of times over the 1000 tests that the $D_{10-NN}$ Shapley values-based counterfactuals yield a counterfactual ability better than the Shapley values computed on $D_{D-TRAIN}$, $D_{D-LAB}$, and $D_{D-PRED}$ (black bars), and vice-versa (green bars). Comparing such figures, it is evident how the $D_{10-NN}$ counterfactuals tend to be much better than those based on $D_{D-TRAIN}$, $D_{D-LAB}$, and $D_{D-PRED}$, and that such superiority is more marked when the proposed $c_M$ distance is adopted (Figure \ref{fig:cm}).

It is worth mentioning that the percentage of times in which one of the five prescription variables have been modified in the induced counterfactuals is $54\%$ for $c_{L_1}$ is $47\%$ for $c_M$, confirming that the latter yields closer, hence better, conterfactuals. \par

% \clearpage
% \begin{table}[h!]
% \centering
% \begin{tabular}{c|c|c|c|c|c|c|c|c|c|c|c|c|}
% \hline
%  & P\_acc & P\_filt &P\_eff & P\_ret &Q\_bp  &Q\_rep & Q\_dial &Q\_pbp & Q\_pfr& $\Delta$P& TMP&TMPa \\ \hline
% $c_{L_1}$ & 689 & 304 & 204 &407 & 305 & 601 & 648 & 491 &662 & 657 & 185 & 186  \\
% $c_M$ &540&218&251&416&275&680&526&566&307&349&150&87 \\
% \hline
% \end{tabular}
% \caption{Filter codes and their TMP filter values.}
% \label{tab:---}
% \end{table}
% \clearpage
 \section{Conclusions}\label{sec:conclusions}

The proposed experimental study suggests that data collected in the CRRT can be exploited to train  machine learning models to accurately predict membrane fouling events at different forecasting horizons, in a straightforward tabular data framework.
%In conclusion, this experimental study has provided evidence regarding the feasibility of using ensemble models for predicting membrane fouling events during CRRT. 
The integration of the ADASYN methodology effectively mitigated the class imbalance, enhancing the performance of the models. Moreover, a post-processing filtering of the prediction allow to avoid oscillating phenomena and to better detect true positive cases. The performance obtained showed to be robust with respect to different length forecasting horizons. Furthermore, the application of the SHAP method highlighted the most significant predictor variables, guiding the study to a feature reduction methodology that potentially enhances model interpretability without sacrificing accuracy. Shapley values are also exploited to determine good quality counterfactual explanations, especially when the proposed Mahalanobis metric is used. Such counterfactuals seem to be promising guidelines to avoid membrane fouling events. 
The study established a proof of concept for a data-driven approach leveraging machine learning techniques to predict and try to avoid blocking events during CRRT. The practical implications for clinicians and nurses managing CRRT are significant  as it could be used as an auxiliary tool to manage in advance a potential membrane fouling event so as to mitigate its impact. \par 
The experiments are carried out on historical data and this may constitute a limitation of the study.  Therefore future extension may concern the application of the proposed methodology in-vivo or in a simulated environment.
\section{Funding acknowledgments}
This project has been funded by PRIN 2022 - Precision medicine in renal replacement therapy for critically ill patients - CUP Master B53D2302084 0006.

\bibliographystyle{plain} 
\bibliography{biblio}
\vspace{12pt}
\color{red}
%IEEE conference templates contain guidance text for composing and formatting conference papers. Please ensure that all template text is removed from your conference paper prior to submission to the conference. Failure to remove the template text from your paper may result in your paper not being published.

\end{document}